\documentclass[10pt]{article}
\usepackage[margin=1in]{geometry}
\usepackage{amsmath,amssymb,amsthm}
\usepackage{lmodern}
\usepackage[round]{natbib}
\usepackage{graphicx}
\usepackage{booktabs}
\usepackage{placeins}
\usepackage[expansion=false]{microtype}
\usepackage[colorlinks=true,linkcolor=blue,citecolor=blue,urlcolor=blue]{hyperref}
\hypersetup{pdfauthor={Peng Xie, Amr Alanwar},pdftitle={The Intruder Threshold: A Spectral Law for LoRA Fine-Tuning}}
\usepackage{xcolor}
\newtheorem{proposition}{Proposition}
\newcommand{\R}{\mathbb{R}}

\title{The Intruder Threshold:\\ A Spectral Law for LoRA Fine-Tuning}
\author{Peng~Xie and Amr~Alanwar%
\thanks{Peng Xie and Amr Alanwar are with the TUM School of Computation, Information and Technology, Department of Computer Engineering, Technical University of Munich, 74076 Heilbronn, Germany (e-mail: p.xie@tum.de; alanwar@tum.de).}}
\date{}

\begin{document}
\maketitle

\begin{abstract}
LoRA fine-tuning can create intruder dimensions: new leading singular
vectors of the updated weight matrix $W+BA$ that are nearly orthogonal to
all pretrained singular vectors and that drive catastrophic forgetting.
Since their discovery, no theory has predicted, layer by layer on measured
spectra, when they appear. We derive
a per-layer critical update strength
$s^\ast=\bar\theta/(\gamma\sigma_1(BA))$, computed from the measured
spectrum of $W$ alone through the rectangular spiked-deformation transform,
together with an exact secular-equation characterization of the updated
spectrum, with no fitted parameters. In a pre-specified study spanning four dense Transformer families, a
state-space model, a mixture-of-experts model, and an encoder-decoder
(18 adapters, 9{,}840 layer scans), the law localizes the empirical
threshold within a factor of two on $82\%$ of layers, separates
intruder-bearing from intruder-free layers at deployment with a mean AUC of
$0.89$, holds unchanged on six third-party adapters, and predicts where
WikiText-2 perplexity begins to degrade; a combination of the two pre-specified edge evaluations reaches $98\%$ and
is confirmed out-of-bag on the external adapters ($0.997$). Full fine-tuning disperses
its update far below the threshold of every layer, which resolves the
asymmetry between LoRA and full fine-tuning. Norm-matched interventions
confirm that threshold-crossing layers, rather than update magnitude, carry
the forgetting, and a spike-budget rule derived from the thresholds, requiring one SVD and
no validation sweeps, reduces forgetting by $62\%$ on the most fragile
model at no task cost.
\end{abstract}

\section{Introduction}
Low-rank adaptation (LoRA) \cite{hu2022lora} updates a pretrained weight
matrix $W\in\R^{n\times m}$ as $W+\gamma BA$ with $B\in\R^{n\times r}$,
$A\in\R^{r\times m}$, $r\ll\min(n,m)$. \citet{shuttleworth2024lora} showed
that LoRA-tuned matrices contain \emph{intruder dimensions}: top singular
vectors $y_j$ of the tuned matrix with
$\max_i|\langle y_j,u_i\rangle|<\tau$ against \emph{all} pretrained singular
vectors $u_i$; full fine-tuning produces none, and causal intervention on
intruder singular values restores pretraining behavior, so intruders cause
forgetting. The phenomenon is widely replicated and cited, but existing
accounts are qualitative (a toy sketch in their Appendix~A), live in
representation space rather than weight-spectrum space
\cite{koubbi2026forgetting}, or are span-containment arguments that cannot
distinguish LoRA from full fine-tuning \cite{yao2026gain}.

We ask a quantitative question: at what update strength does the intruder
appear? Our answer is a per-layer critical strength computed purely from the
pretrained spectrum, with no training curves, no fitting, and no tunable
constants, validated by a pre-specified falsification protocol on real
7--8B LLMs. Three outcomes support it: the threshold is localized within a
factor of two on $82.2\%$ of layers; the forgetting knee sits where the
median per-layer threshold predicts; and the residual, systematic
$1.2$--$1.5\times$ early arrival of real intruders isolates a measurable
quantity, the anti-alignment bias of trained adapters, which we subsequently
attribute to alignment structure and multi-spike order statistics.

\paragraph{Contributions.}
(1)~A measurement instrument: standard thin-SVD updating
\citep{zha1999,brand2006} arranged so that a $(p{+}r)$-dimensional core
matrix returns the full singular system of $W+s\gamma BA$ and its overlaps
with the pretrained basis exactly (\S\ref{sec:theory}), making $9{,}840$
layer scans affordable; the equivalent secular equation is verified in
float64 on real checkpoints to about $10^{-3}$ relative error.
(2)~A universal emergence law: the critical spike strength
$\bar\theta = D_{\mu}(z_0)^{-1/2}$ evaluated on the measured spectrum
$\mu$ of $W$, giving $s^\ast=\bar\theta/(\gamma\sigma_1(BA))$, the outlier
trajectory $\sigma_{\mathrm{out}}(s)$, and overlap decay, all with no
fitted parameters (\S\ref{sec:theory}).
(3)~A pre-specified validation with go/no-go gates and falsification bars
fixed before the runs (\S\ref{sec:exp}); all bars were met, and the single
failed subclaim, non-vacuous Wedin rotation control, is itself an
informative result about real LLM spectra.
(4)~The anti-alignment bias: real adapters produce intruders
$1.2$--$1.5\times$ earlier than the free-position prediction, consistently
in sign on 22 of 24 adapters including the external ones
(\S\ref{sec:bias}).
(5)~Prediction converts to control (\S\ref{sec:interv}): norm-matched causal
interventions show predicted-supercritical layers carry up to $240\times$
more forgetting than subcritical ones; a five-cell equal-norm surgery isolates
the sufficient statistic; and a projection-free per-layer budget rule cuts
forgetting $62\%$ on the most fragile model at no task cost.

\section{Theory: an exactness ladder}\label{sec:theory}
Let $W=U\Sigma V^\top$ be the reduced SVD of a pretrained weight matrix,
with $p=\min(n,m)$ singular values $\sigma_1\ge\cdots\ge\sigma_p$ and
aspect ratio $c=p/\max(n,m)$, and let the deployed update be
$\Delta(s)=s\gamma BA$ with LoRA scaling $\gamma=\alpha/r$ (or
$\alpha/\sqrt r$ for rsLoRA \cite{kalajdzievski2023rslora}). The theory
forms a ladder. The exact rung characterizes the updated spectrum with no
assumptions and serves as the measurement instrument. The universal rung
assumes only that the update sits in generic position relative to the
pretrained basis and yields a prediction with no fitted parameters; this is
the hypothesis under test. The intermediate rung replaces parts of that
assumption with measurements of the trained adapter.

\subsection{The exact rung: core matrix and secular equation}
Write $P_B=U^\top B$ and $P_A=V^\top A^\top$, and let $Q R$ be the QR
factorization of the component of the update outside the span of the
incomplete side of the basis. Then the following holds.
\begin{proposition}\label{prop:core}
The nonzero singular values of $W+s\gamma BA$ equal those of the
$(p{+}r_b)\times(p{+}r_a)$ core matrix
$\;C(s)=\begin{bmatrix}\Sigma&0\\0&0\end{bmatrix}
+ s\gamma\,\begin{bmatrix}P_B\\R_B\end{bmatrix}
\begin{bmatrix}P_A\\R_A\end{bmatrix}^{\!\top}$,
where exactly one of $r_b,r_a$ is nonzero ($r$ on the incomplete side, $0$
on the complete side). Moreover, if $C=U_c\Sigma_c V_c^\top$, then row
$i\le p$ of $U_c$ (resp.\ $V_c$) is exactly the overlap of the $j$-th new
left (right) singular vector with the pretrained $u_i$ ($v_i$).
\end{proposition}
The proof is an elementary orthonormal completion argument, but the
consequence is practical: the entire intruder diagnostic (top singular
values, the Shuttleworth max-cosine criterion against all pretrained
directions, and the subspace overlap order parameters $m_K(s)$) is computed
exactly from one small-matrix SVD per $(\mathrm{layer},s)$ pair, with no
large-matrix SVD per scan point and no randomized approximation.

To connect this algebra with spiked-matrix theory, embed $W$ in the
symmetric matrix
$H_0=\big(\begin{smallmatrix}0&W\\W^\top&0\end{smallmatrix}\big)$,
whose eigenpairs are $\pm\sigma_i$ with eigenvectors
$(u_i,\pm v_i)/\sqrt2$, together with $|n-m|$ zero modes. A rank-$r$
update of $W$ becomes a symmetric update of rank at most $2r$, and a
standard determinant identity states that a value
$\lambda\notin\mathrm{spec}(H_0)$ belongs to the updated spectrum exactly
when $\det\big(J-K^\top(\lambda-H_0)^{-1}K\big)=0$, where $K$ collects
the update factors and $J$ swaps its two blocks \cite{bunch1978rank}.
Expanding the resolvent in the eigenbasis of $H_0$ reduces every entry of
this determinant to the weights $\lambda/(\lambda^2-\sigma_i^2)$,
$\sigma_i/(\lambda^2-\sigma_i^2)$, and $1/\lambda$. For a rank-one update
$\theta uv^\top$ the determinant is two by two and the condition reads
\begin{equation}\label{eq:secular}
\Big(X(\lambda)-\frac1\theta\Big)^{2}
=\Phi_u(\lambda)\,\Phi_v(\lambda),
\end{equation}
where
\begin{equation}\label{eq:overlapsums}
\Phi_u(\lambda)=\sum_i\langle u,u_i\rangle^2
\frac{\lambda}{\lambda^2-\sigma_i^2},\qquad
\Phi_v(\lambda)=\sum_i\langle v,v_i\rangle^2
\frac{\lambda}{\lambda^2-\sigma_i^2}+\frac{\|v_\perp\|^2}{\lambda},\qquad
X(\lambda)=\sum_i\langle u,u_i\rangle\langle v,v_i\rangle
\frac{\sigma_i}{\lambda^2-\sigma_i^2},
\end{equation}
and $v_\perp$ is the component of $v$ in the null space of $W$; we adopt
the convention $n\le m$ (transpose otherwise), so the null-space term
appears on the right side only. Equation~\eqref{eq:secular} is exact: given the overlaps of
$(u,v)$ with the pretrained basis, it determines every outlier singular
value of $W+\theta uv^\top$. The later rungs are controlled limits of this
equation. We also verify it directly in float64 on real checkpoints, where
the outliers of the fp32 scan satisfy it to a maximum relative error below
$10^{-3}$, and the bias dissection of \S\ref{sec:bias} operates on its
terms.

\subsection{The universal rung: the free-position limit}
Suppose the spike directions are generic. When $u$ is a uniformly random
unit vector its squared overlaps concentrate at their mean,
$\langle u,u_i\rangle^2\approx1/n$, and likewise for $v$, while the
products $\langle u,u_i\rangle\langle v,v_i\rangle$ carry independent
signs and average to zero. Under these three replacements the sums in
Eq.~\eqref{eq:overlapsums} lose all dependence on the directions:
$\Phi_u\to\varphi$, $\Phi_v\to c\varphi+(1-c)/\lambda$, and $X\to0$,
with
\begin{equation}\label{eq:D}
\varphi(z)=\frac1p\sum_i\frac{z}{z^2-\sigma_i^2},\qquad
D(z)=\varphi(z)\Big[c\,\varphi(z)+\frac{1-c}{z}\Big].
\end{equation}
Equation~\eqref{eq:secular} then collapses to $D(\lambda)=1/\theta^2$,
which is the rectangular spiked-deformation condition of
\citet{benaych2012singular}, evaluated here on the empirical measure of the
actual layer rather than on a limiting distribution. Since $D$ decreases
monotonically on $(\sigma_1,\infty)$ and tends to zero, a solution exists
exactly when $\theta$ exceeds a critical strength, and for a LoRA update
whose leading spike has strength $s\gamma\sigma_1(BA)$ this criterion
becomes a per-layer critical scale:
\begin{equation}\label{eq:law}
\boxed{\;s^\ast=\bar\theta\,/\,\big(\gamma\,\sigma_1(BA)\big),
\qquad \bar\theta=D(z_0)^{-1/2}\;}
\end{equation}
where $z_0$ is an evaluation point just above the spectrum, discussed next.
Above threshold the outlier location solves
$D(\sigma_{\mathrm{out}}(s))=(s\gamma\sigma_1(BA))^{-2}$, and the
limiting overlap of the new singular vector with the spike direction is
$-2\varphi(\rho)/(\theta^2D'(\rho))$ on the smaller side. Every quantity
derives from one SVD of $W$.

\subsection{The evaluation point and its sensitivity}\label{sec:eps}
The empirical measure is discrete, so $\varphi$ has a pole at every
$\sigma_i$: the $i=1$ term behaves as
$\sigma_1/\big((z-\sigma_1)(z+\sigma_1)\big)$ and diverges as
$z\to\sigma_1^{+}$. Evaluating $D$ exactly at the top of the spectrum
would therefore return $\bar\theta=0$ for any model, and the evaluation
point must sit slightly outside, $z_0=\sigma_1(1+\varepsilon)$, where
$\varepsilon>0$ is a choice rather than a derived quantity. The prediction
is meaningful only if it does not depend materially on this choice, so we
treat $\bar\theta$ as a function of $\varepsilon$ and report, for every
layer, the sensitivity ratio
\begin{equation}\label{eq:sens}
R=\max_{\varepsilon\in E}\bar\theta(\varepsilon)\,\Big/\,
\min_{\varepsilon\in E}\bar\theta(\varepsilon),
\qquad E=[10^{-4},\,3\times10^{-2}],
\end{equation}
which sweeps the evaluation point across more than two orders of magnitude.
A small $R$ means that $\bar\theta(\varepsilon)$ has a plateau: the
threshold is a collective property of the whole spectrum, and any
reasonable evaluation point reads off essentially the same value. A large
$R$ means that the value is dominated by the few singular values closest to
the edge and is an artifact of where one stands. A synthetic example makes
the distinction concrete. For a heavy-tailed spectrum with a sparse top
($p=4096$), $\bar\theta(\varepsilon)$ moves through
$0.53,\,0.76,\,0.90,\,0.96,\,0.99,\,1.01$ across $E$ and settles on a
plateau ($R=1.9$); for a spectrum with half of its mass packed within
$0.5\%$ of the top, the same sweep gives values from $0.007$ to $0.161$
($R=22.7$) and no plateau exists. Gate G2 of the pre-specification declares
the prediction degenerate when the median $R$ across layers exceeds $10$.
The observed medians are $2.2$ to $2.4$ on the five causal base models and
$3.4$ to $4.3$ on the encoder-decoder and mixture-of-experts models, in all
cases far from the degenerate regime, and this stability was not guaranteed
in advance. A
second, related choice is which edge to stand above: the full edge uses
$\sigma_1$ itself, while the bulk edge first trims isolated top outliers by
a gap test and stands above the remaining bulk. Both variants are
pre-specified and stored per layer; the primary results use the full edge,
and \S\ref{sec:exp} analyzes the difference between them. Because of these
conventions we describe the law as having no fitted parameters rather than
no free parameters: $\varepsilon$ and the edge are declared choices,
constrained by the sensitivity analysis, not quantities tuned to data.

\subsection{The intermediate rung: alignment correction}
Replacing $\sigma_1(BA)$ by $\sigma_1(P^\perp_K\,BA\,P^\perp_K)$, the
spike strength in the orthogonal complement of the top-$K$ pretrained
subspace, gives an alignment-corrected prediction ($K\in\{16,64,256\}$
reported). Where a real adapter falls on this ladder is itself a
measurement of how far trained updates are from free position.

\section{Pre-registered validation}\label{sec:exp}
\paragraph{Protocol.}
All gates and bars were fixed in the analysis code before any scan ran; we
use pre-specified in this internal sense throughout, as no external registry
was used. \emph{Census}: full-layer SVD
census of Qwen2.5-7B, Llama-3.1-8B, Mistral-7B-v0.3, OLMo-2-7B (868 block
matrices); gate G2 terminates the study if $\bar\theta$ is degenerate
(non-finite, no $\varepsilon$-plateau, or no cross-layer discrimination).
\emph{Falsifier}: 8 LoRA adapters (2 models $\times$ \{MetaMathQA
\citep{yu2024metamath}, alpaca-cleaned \citep{taori2023alpaca}\}
$\times$ $r\in\{16,256\}$, $\alpha=2r$, 1000 steps); for each
adapted matrix, scan $W+s\gamma BA$ over two decades of $s$ (25 points,
auto-centered; the deployed point $s{=}1$ always included) and record the
exact spectrum and overlap diagnostics of Proposition~\ref{prop:core};
empirical $s^\ast_{\mathrm{emp}}$ is the first grid point with an intruder
under the original criterion ($\tau=0.5$, top-10; $\tau\in\{0.3,0.7\}$ stored
for sensitivity). Bars: the universal law passes if $\ge70\%$ of evaluable
layers satisfy $|\log_2 (s^\ast_{\mathrm{pred}}/s^\ast_{\mathrm{emp}})|\le 1$
and Spearman $\rho\ge0.6$. Throughout the paper, hit@$\times2$ denotes the
fraction of evaluable layers whose prediction lands within a factor of two
of the measurement in this sense; $\rho$ denotes the Spearman rank
correlation between predicted and empirical thresholds across the layers of
one adapter; pooled values average the per-adapter results; and a layer is
evaluable when both quantities are defined, which holds for every scanned
layer in our runs.

\paragraph{Census (gate G2 passed).}
On all four models, $100\%$ of layers have finite $\bar\theta$; the median
sensitivity ratio $R$ of Eq.~\eqref{eq:sens} is $2.2$ to $2.4$ against the
pre-specified degeneracy bound of $10$, and the cross-layer coefficient of variation is
$0.44$--$0.91$: the per-layer threshold is well-defined and discriminative on
real, heavy-tailed LLM spectra \citep{martin2021implicit}, which was not
guaranteed in advance. Extending the census to Mamba-2.8b, OLMoE-1B-7B, and
FLAN-T5-XL preserves both properties, with the sensitivity medians reported
in \S\ref{sec:eps}.

\paragraph{Falsifier: the law holds.}
Pooled over all $9{,}840$ layer-scans of the 18 adapters, hit@$\times2$ is
$82.2\%$ with a bootstrap $95\%$ confidence interval of $[0.785,0.862]$
over adapters, entirely above the pre-specified bar, and $\rho=0.661$ with
interval $[0.583,0.736]$, whose point estimate clears the bar while the
lower bound sits marginally below it. The transition of every layer was
captured within the scan grid, so censoring cannot manufacture hits, and
because the empirical threshold is the first grid point at which an
intruder is present, grid quantization can only delay it by at most one
step ($0.28$ in $\log_2$), which biases against rather than toward the
early-arrival finding.

\begin{figure}[t]\centering
\includegraphics[width=\linewidth]{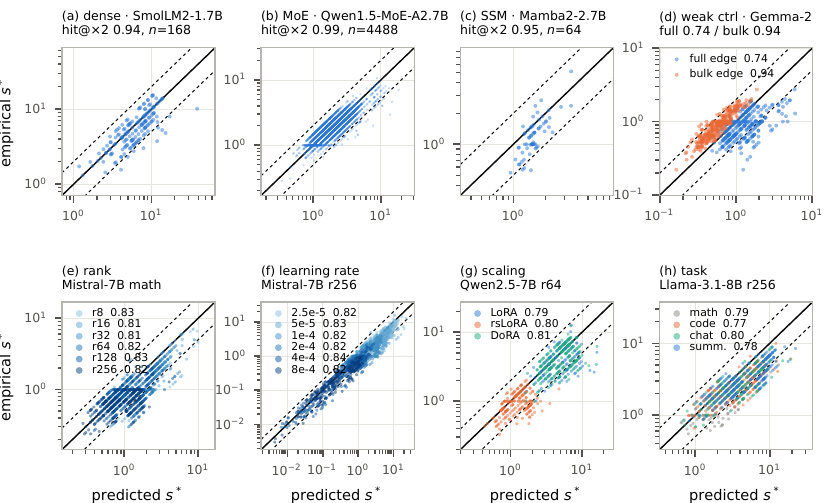}
\caption{Predicted versus empirical per-layer threshold (log-log; solid, exact;
dashed, factor of two). Top: the $r{=}256$ adapter with the highest hit@$\times2$
of each architecture class, and the pre-registered weak control under both
edges. Bottom: the rank, learning-rate, parameterization and task series.}
\label{fig:scatter}
\end{figure}

\paragraph{Calibration against nulls.}
The law supplies three capabilities that no scan-dependent reference can
match: an absolute scale available before any training or scanning, a
per-layer ranking ($\rho=0.66$ under the full edge, $0.90$ out-of-bag
under the bracketed edge), and layer-level classification of deployment
risk, separating intruder-bearing from intruder-free layers at $s{=}1$ with
a mean ROC-AUC of $0.89$ across the 14 adapters that contain both classes.
Two reference predictors quantify how much of the localization metric these
capabilities explain. An oracle constant that assigns every layer the
median empirical threshold of its own adapter reaches hit@$\times2$
$=0.83$, since thresholds cluster within an adapter; it requires running
the scans it claims to predict, carries no ranking information, and scores
AUC $0.5$ on the deployment classification by construction. The toy
condition of \citet{shuttleworth2024lora}, which replaces $\bar\theta$ by
$\sigma_{\max}$, reaches $0.71$ with $\rho=0.72$. Localization within a
factor of two is therefore the weakest output of the law rather than its
content; the binding evidence is the ranking, the a-priori scale, the
deployment AUC, and the $0.98$ localization of the bracketed edge. Per-adapter binomial confidence half-widths on
hit@$\times2$ are at most $0.066$. Training-seed variance is also small:
retraining the Mistral and Qwen $r256$ configurations with two further
seeds moves hit@$\times2$ by at most $0.02$ and $\rho$ by at most $0.06$,
the median per-layer cross-seed range of the empirical threshold equals the
scan-grid step ($0.26$--$0.27$ compared with $0.28$ in $\log_2$), and the
layer ranking agrees between seeds at $\rho=0.88$--$0.93$. The per-layer
threshold is therefore a reproducible property of the configuration rather
than an artifact of a single run. Table~\ref{tab:main} and Figure~\ref{fig:scatter} give per-adapter
results. At the deployed strength $s{=}1$, the fraction of layers carrying
intruders in the $r{=}256$ adapters spans $0\%$ (OLMo-2), $8$--$17\%$
(Qwen, Llama), $45\%$ (Mamba), and $91\%$ (Mistral), an
order-of-magnitude cross-model spread that mirrors the per-layer thresholds
themselves and qualitatively reproduces \citet{shuttleworth2024lora}.

\begin{table}[t]\centering\small
\begin{tabular}{lccc}
\toprule
Adapter & hit@$\times2$ & Spearman $\rho$ & med.\ $|\log_2|$ err \\
\midrule
Llama-3.1-8B instr $r16$ / $r256$ & 0.82 / 0.79 & 0.76 / 0.80 & 0.55 \\
Llama-3.1-8B math $r16$ / $r256$  & 0.80 / 0.79 & 0.82 / 0.83 & 0.55 \\
Qwen2.5-7B instr $r16$ / $r256$   & 0.82 / 0.83 & 0.50 / 0.66 & 0.28 \\
Qwen2.5-7B math $r16$ / $r256$    & 0.80 / 0.82 & 0.45 / 0.58 & 0.28 \\
Mistral-7B math $r16$ / $r256$    & 0.81 / 0.82 & 0.69 / 0.55 & 0.55 \\
OLMo-2-7B math $r16$ / $r256$     & 0.71 / 0.70 & 0.33 / 0.39 & 0.55 \\
Mamba-2.8b math $r16$ / $r256$    & 0.92 / 0.94 & 0.49 / 0.76 & 0.55 \\
OLMoE-1B-7B math $r16$ / $r256$   & \textbf{0.99} / \textbf{0.98} & 0.84 / 0.86 & 0.28 \\
FLAN-T5-XL math $r16$ / $r256$    & 0.74 / 0.72 & 0.82 / 0.77 & 0.55 / 0.28 \\
\midrule
pooled (18 adapters, 9{,}840 scans) & \textbf{0.822} & \textbf{0.661} & --- \\
\bottomrule
\end{tabular}
\caption{Universal (free-position) prediction vs.\ empirical intruder
threshold, per adapter. Pre-registered bars: $0.70$ / $0.60$. Median $|\log_2|$ errors are
quantized by the scan-grid resolution ($0.28$ per grid step). The low
per-family $\rho$ of OLMo-2 and of the weak Qwen adapters is diagnosed in
the edge-variant analysis: it reflects the conservative
$\sigma_{\max}$-edge evaluation on module types with outlier-heavy spectra
rather than the law itself; the bulk-edge variant restores
$\rho=0.82$--$0.96$ on every family.}
\label{tab:main}
\end{table}

\paragraph{Edge-variant analysis.}
Both pre-specified edges of \S\ref{sec:eps} were stored per layer:
the conservative full edge (primary, all results above) and the trimmed
bulk edge. The full-edge variant is
systematically over-conservative on module types whose spectra carry large
isolated outliers, worst for \texttt{o\_proj} in OLMo-2
($\sigma_{\max}/\mathrm{bulk}=3.7$; median emp/pred $0.32$; within-type
$\rho<0$), and this is what depresses the pooled rank correlation of the
low-$\rho$ families, whose within-type rankings remain sound. The bulk-edge
variant repairs the ordering on every family ($\rho=0.82$--$0.96$, mean
$0.90$; OLMo-2 $0.33\to0.82$) at the cost of an early bias (median ratio
$1.5$). The two variants bracket the effective edge, and their geometric
mean, a parameter-free combination of two pre-specified estimators, is
nearly unbiased: mean hit@$\times2$ $=0.98$ (13 of 18 adapters at $1.00$),
mean $\rho=0.83$, median ratio $1.06$. The combination was chosen on our
own adapters; on the six external adapters, which played no role in that
choice, it achieves out-of-bag hit@$\times2$ $=0.997$ with mean
$\rho=0.90$. The pre-specified primary endpoints stand as reported above.

\begin{figure}[t]\centering
\includegraphics[width=0.94\linewidth]{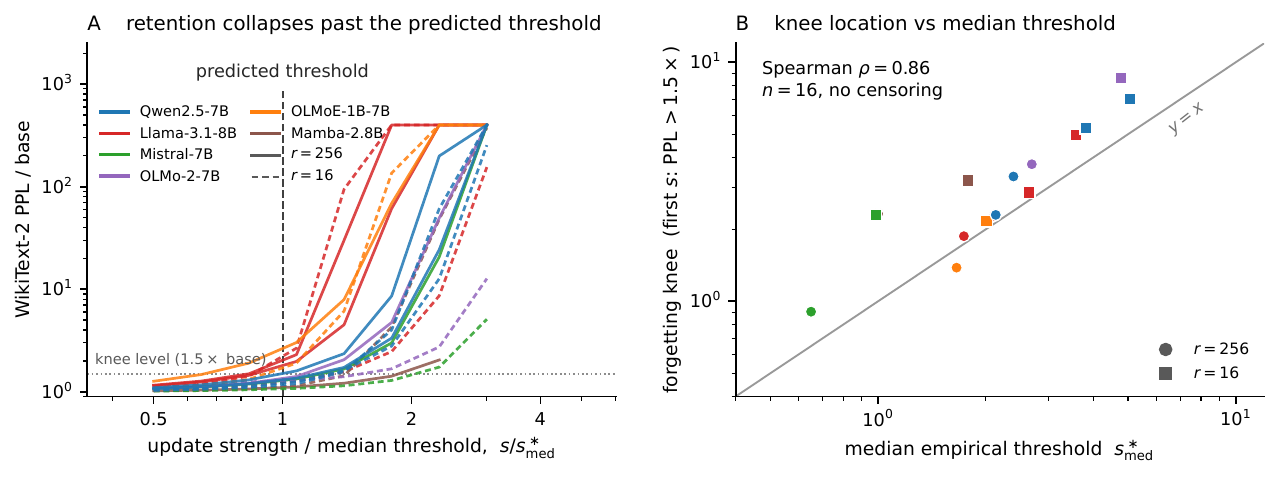}
\caption{Forgetting begins where the spectrum says it will. Left: WikiText-2
perplexity relative to base for all 16 probed adapters, with update strength
rescaled by the median empirical threshold of each adapter; retention is intact
left of the line and collapses beyond it. Right: knee location (first strength
at which perplexity exceeds $1.5\times$ base) against the median threshold
with the identity line; the dense per-adapter grids leave no censored
adapters (Spearman $0.86$, $n=16$).}
\label{fig:knee}
\end{figure}

\paragraph{The forgetting knee is where the law says it is.}
Rebuilding $W+s\gamma BA$ in the full model and measuring WikiText-2
perplexity: adapters with a low median threshold degrade early, while those
with a high threshold remain flat through the probed range. For example,
Llama math $r256$ (median $s^\ast_{\mathrm{emp}}=1.74$) has PPL
$6.3\to7.2\to13.7\to15{,}562$ at $s=0.5,1,2,4$, whereas Qwen instr $r16$
(median $3.80$) stays at $6.6\to6.7\to6.8\to7.8$. Across the 16 probed adapters
the PPL knee tracks the median threshold over a $7.8\times$ range. Mistral
$r256$ is already past threshold at deployment (median
$s^\ast_{\mathrm{emp}}=0.65<1$; $91\%$ of layers carry intruders; PPL
$9.7\to9{,}092$ between $s{=}1$ and $s{=}2$), while OLMo-2 $r16$ sits far
below it (median $4.78$; no deployed intruders; PPL flat, $6.0\to7.5$,
through $s{=}4$), so the two additional families bracket the earlier results
as their spectra predict. Defining the knee as the first strength at which
WikiText-2 perplexity exceeds $1.5\times$ its base value, and probing each
adapter on a geometric strength grid spanning $0.5$--$3\times$ its median
empirical threshold, all
16 knees fall inside the probed range and correlate with the median
empirical threshold at Spearman $0.86$ ($p=1.8\times10^{-5}$;
Figure~\ref{fig:knee}). The correlation is insensitive to the knee
definition ($\rho=0.86$ at $1.2\times$ base, $0.84$ at $2.0\times$), and
the knee sits at $0.8$--$2.3\times$ the median threshold (median
$1.4\times$): intruders appear first, and measurable retention loss follows
within a factor of about two in strength.
Task perplexity degrades only well above the knee.

\paragraph{Intruders are not an attention phenomenon.}
The law never references attention, so we tested it where attention does not
exist: LoRA ($r\in\{16,256\}$, math) on the input projections of
Mamba-2.8b, a selective state-space model. Intruder dimensions appear in
$45\%$ of layers at deployment for $r256$, to our knowledge their first
documentation in an SSM. The unmodified formula localizes their emergence
with the best per-adapter accuracy of the study
(hit@$\times2$ $=0.92/0.94$, $\rho=0.49/0.76$), the forgetting knee again
tracks the median threshold (PPL $9.8\to14.2\to2{,}478$ for $s=1,2,4$ at
median $s^\ast_{\mathrm{emp}}=1.00$), the anti-alignment bias has the same
sign and magnitude ($0.68$--$0.83\times$), and the float64 secular check
passes ($\le6.4\times10^{-4}$). Two further architecture classes complete
the picture. On OLMoE-1B-7B, a mixture-of-experts model, we adapt attention
and all 64 experts per layer (3{,}136 matrices per adapter) and observe the
best accuracy of the study: hit@$\times2$ $=0.99/0.98$ with
$\rho=0.84/0.86$, intruders in $9\%$ and $19\%$ of matrices at deployment,
and a forgetting knee at the predicted location. The census also shows that
the 64 experts of a single layer are spectrally heterogeneous, with
$\bar\theta$ spanning a $2.6\times$ range (coefficient of variation
$0.21$) within one layer, so per-expert budgets are meaningful. On
FLAN-T5-XL, an encoder-decoder model, the full-edge variant is the weakest
of the study ($0.74/0.72$, consistent with its high edge sensitivity),
while the geometric-mean edge restores $0.88/0.91$ with $\rho\approx0.88$
and median ratio $1.08$; the encoder-decoder is also the most resistant
architecture at deployment ($1$--$4\%$ intruder-carrying matrices).
Architecture independence is therefore a measured property rather than a
design assumption.

\begin{figure}[t]\centering
\includegraphics[width=0.72\linewidth]{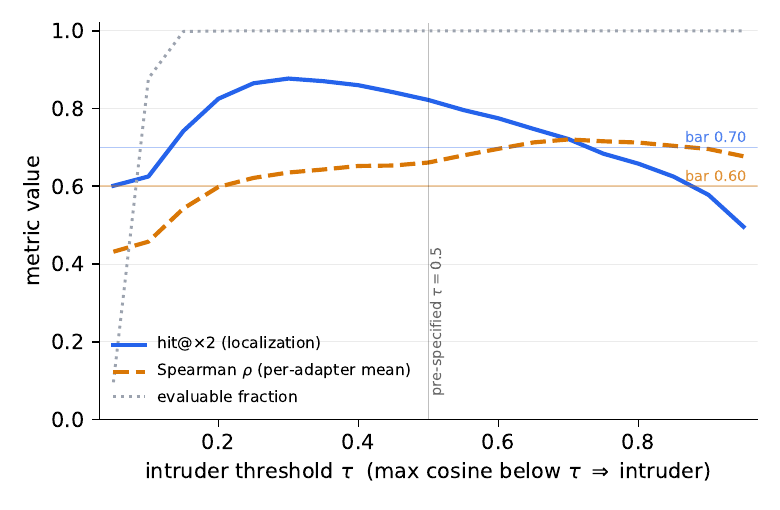}
\caption{Both pre-specified bars hold for every $\tau\in[0.25,0.70]$; the
pre-specified $\tau{=}0.5$ sits inside the window (all 9{,}840 scans;
dotted: evaluable fraction).}
\label{fig:tau}
\end{figure}

\begin{figure}[!t]\centering
\includegraphics[width=0.9\linewidth]{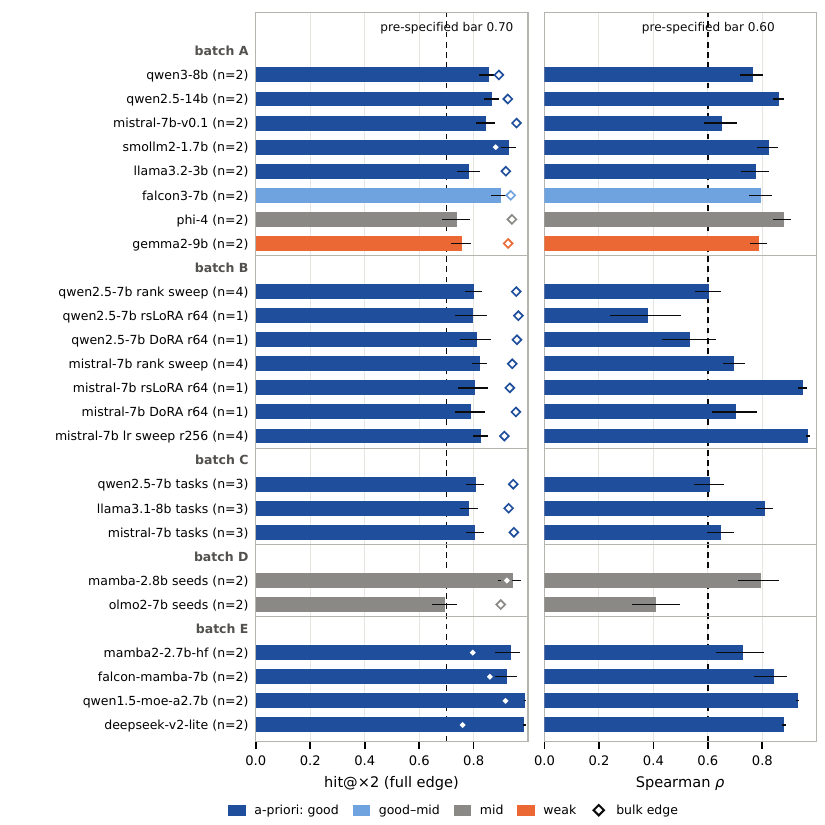}
\caption{Second batch, pooled per family or configuration: hit@$\times2$ (bars, full
edge; diamonds, bulk edge) and Spearman $\rho$, coloured by the pre-registered
label, with the pre-specified bars (dashed) and 95\% intervals.}
\label{fig:prereg}
\end{figure}

\paragraph{A second pre-specified batch.} After the study above we froze a
second unit list (59 units: eight further base models from 1.7B to 14B
including a post-trained checkpoint, a rank and learning-rate sweep, rsLoRA
and DoRA, three further tasks, seed replicates, six external adapters, two
state-space and two mixture-of-experts models), each unit labelled with an
expected outcome before submission (Appendix~\ref{app:ext}). Figure~\ref{fig:prereg}
puts every family against its label: the batches pool to hit@$\times2$ of
$0.84$ (new families), $0.81$ (rank, learning rate, parameterization), $0.80$
(tasks) and $0.96$ (SSM and MoE), all above the bars; the families labelled
weak or intermediate (Gemma-2-9B, OLMo-2, Phi-4) sit at or below the
full-edge bar and are repaired by the bulk edge (Gemma-2-9B: $0.74$ under
the full edge, $0.94$ under the bulk edge), and one label was too cautious
(Phi-4, $\rho$ of $0.86$). The pooled $\rho$ of a sweep exceeds that of its
members because the sweep spans thresholds across a decade. Along the knob
series of Figure~\ref{fig:scatter}, rank ($r{=}8$ to $256$) and learning rate
($2.5\times10^{-5}$ to $8\times10^{-4}$) slide the thresholds down the
diagonal by more than a decade while localization stays at $0.81$--$0.84$,
the scaling convention moves the same $r{=}64$ adapter across the plane, and
four tasks of the same base overlap; the consistent below-diagonal shift is
the anti-alignment bias of \S\ref{sec:bias}.

\section{From prediction to intervention and control}\label{sec:interv}
\paragraph{Norm-matched causal intervention.}
The law identifies the responsible layers before any intervention is run.
Scaling only the predicted supercritical layers of an adapter versus only
the subcritical ones, with the total injected Frobenius norm matched,
separates the amount of update from where it lands relative to threshold.
On Mistral $r256$ ($221$ supercritical layers at $s{=}2$), the
supercritical-only condition drives WikiText-2 PPL from $4.9$ to $9{,}340$,
while the norm-matched subcritical condition reaches only $39$, a
$240\times$ asymmetry at equal injected norm. On Qwen at $s{=}4$ (a
$143$/$53$ split) the matched asymmetry is $1.56\times$ ($+4.20$ versus
$+2.69$ PPL), and on Llama about $1.5\times$; the differential grows with
the supercritical margin, as the law predicts. These ratios are conservative
lower bounds, because norm-matching forces the subcritical set past some of
its own thresholds.

\paragraph{Surgery: the sufficient statistic, isolated.}
Five synthetic equal-norm updates ($\|\Delta\|_F=2\bar\theta$ per layer,
$\sim$24 Qwen matrices) realize the $2\times2$(+1) design. Concentrated
$\times$ generic position: intruders on $100\%$ of layers at median
$s=0.500$ against a predicted $0.5$. Concentrated $\times$ top-aligned and
diffuse $\times$ top-aligned: no intruders at any strength. Concentrated
$\times$ bulk-aligned: a new outlier emerges at $s\approx0.50$ yet is never
an intruder, so spectral escape and novelty are different events, and the
criterion distinguishes them. Diffuse $\times$ generic ($r{=}16$,
per-direction $\theta/\sqrt r$): intruders arrive $3\times$ later (the
naive independent-spike prediction is $4\times$; multi-spike order
statistics account for the gap). Intruder formation is governed by a single
statistic, supercritical strength in generic position, which simultaneously
explains the immunity of full fine-tuning (diffusion) and of aligned LoRA
updates.

\begin{figure}[!t]\centering
\includegraphics[width=\linewidth]{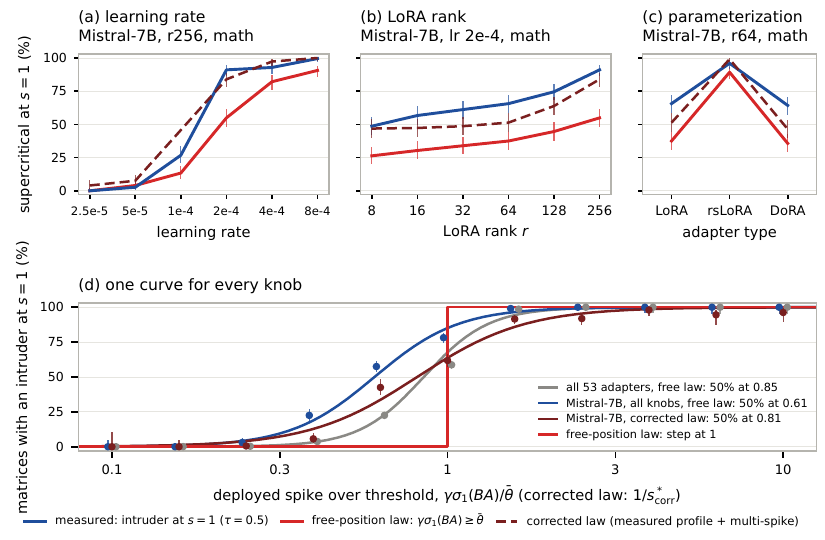}
\caption{The law reads the knobs. (a--c) Mistral-7B matrices with an intruder at
deployment: measured (blue), free-position law (red), law with the corrections
of \S\ref{sec:bias} (dark red, dashed). (d) The same event against the deployed
spike over the threshold, pooled over adapters; the free law is the step at $1$.}
\label{fig:dose}
\end{figure}

\paragraph{Spike-budgeted LoRA.}
This application demonstrates a consequence of the law rather than a new
forgetting-mitigation method; the rule competes on zero overhead, not on
state-of-the-art mitigation. The threshold converts into a per-layer
hyperparameter rule with no
projections and no training-loop changes: cap the per-layer
\texttt{lora\_alpha} so that the end-of-training spike sits at
$\beta\,\bar\theta_\ell$ (PEFT \texttt{alpha\_pattern}; one calibration
constant from a prior run; cap semantics, meaning the rule never raises a
layer toward its threshold). On the most fragile model (Mistral $r256$, with
$91\%$ of layers carrying intruders at deployment), $\beta{=}0.7$ reduces
forgetting from $+4.71$ to $+1.77$ WikiText-2 PPL ($-62\%$) while improving
task perplexity ($1.427\to1.333$). Accuracy metrics agree: the baseline
adapter costs $9.2$ HellaSwag points of retention ($0.626\to0.534$) at
$64.6\%$ GSM8K, and the theory-scaled variants recover about half of the
retention loss ($0.578$--$0.585$) with task accuracy at least maintained
($66.0$--$68.4\%$ against $64.6\%$ at $n{=}500$, within binomial noise). A uniform-$\alpha$ control matched to the
mean strength of the budget rule performs equivalently within noise
(perplexity shows a $16\%$ advantage for per-layer allocation; the accuracy
metrics do not resolve it), so the decisive and robust gain is the global
scale set by the theory: the spectra determine how strong a fine-tune the
model tolerates. A learning-rate control makes the same point through a
different knob: retraining the Mistral adapter at half the learning rate
reduces forgetting to $+1.12$ PPL with task metrics matching the capped
variants ($1.318$ task perplexity, $68.8\%$ GSM8K, $0.590$ HellaSwag). Figure~\ref{fig:dose} extends it to the whole dial: over six learning
rates, six ranks and three parameterizations of Mistral-7B, the fraction of
matrices the law predicts to be supercritical at deployment rises with the
measured fraction, monotone in both knobs. The threshold $\bar\theta$ does not
depend on the knob; the deployed spike $\gamma\sigma_1(BA)$ does, from $0.72$
to $1.14$ of the median threshold across ranks, and rsLoRA at $\alpha{=}2r$
multiplies it by $\sqrt{r}/2$. The free-position prediction sits below the
measurement by a nearly constant margin, which is the anti-alignment bias of
\S\ref{sec:bias}: against the one variable the law uses, the deployed spike
over the layer's threshold (Figure~\ref{fig:dose}d), the measured transition
is a logistic in the log ratio with midpoint $0.61$ over the 19 Mistral
adapters ($3{,}584$ matrices; $0.85$ over all 53 adapters, $29{,}170$
matrices) where the free law puts a step at $1$. With the two computable
corrections of \S\ref{sec:bias} (measured energy profile of each spike over
the full pretrained basis, first crossing among the spikes) the predicted
curves of panels (a--c) close about two thirds of the gap (rank series: free
law $26$--$55\%$, corrected $47$--$84\%$, measured $49$--$91\%$), the median
threshold ratio moves from $0.83$ to $1.05$ and the midpoint of panel (d) to
$0.81$ (12 adapters, $2{,}688$ matrices; the $10^{-4}$ adapter of the main
study has no stored weights). The remainder is not truncation: 64 spikes in
place of 8 change no layer, and the full basis in place of the census top-512
moves the fractions by $1$--$3$ points. It is the independent-spike,
escape-equals-intruder approximation itself; the exact core matrix of
Proposition~\ref{prop:core} couples every spike and reproduces the measurement by
construction.
Scanning the retrained adapter shows why: halving the rate moves the
deployment point from $91\%$ supercritical layers to $27\%$ (median
threshold $0.65\to1.49$), below threshold on the same yardstick that
prices $\alpha$, so the law reads any knob that shrinks the
end-of-training update. Its contribution is the calibration rather than
the choice of knob: it states before training that deployment strength
sits at $1.5\times$ the median threshold for Mistral $r256$ and below
half of it for Qwen, and it prices the required shrinkage per layer. The
Qwen side of that statement is directly testable: the same blind halving
applied to Qwen buys nothing, with forgetting already at floor
($+0.35\to+0.14$ PPL), task perplexity identical ($1.206$), and accuracy
unchanged within noise (GSM8K $83.8\to83.0\%$, HellaSwag
$0.598\to0.596$), exactly as a deployment point below half threshold
predicts. The calibration thus tells you in advance which retraining runs
are worth paying for. A validation sweep over $\alpha$ or the learning
rate could locate a similar operating point at the cost of several extra
training runs per model; the rule provides it from one SVD with none. Dose responses confirm the mechanism in both directions:
$\beta{=}1.3$ (controlled crossing) worsens forgetting by $30\%$, and a
target-mode ablation that raises subcritical layers toward threshold
worsens an otherwise safe model (Qwen: $+0.35\to+0.99$). On Qwen, which is
already safe, the cap rule is correctly inert ($+0.34$ versus $+0.35$
forgetting; GSM8K $82.4$ versus $83.8\%$; HellaSwag identical at $0.598$):
the rule yields large gains where the spectra indicate fragility and does no
harm where they do not. This upgrades the link between intruders and
forgetting from post-hoc correlation \cite{shuttleworth2024lora} to
predicted location.

\paragraph{Practical uses of the law.}
The experiments support four concrete uses. First, a fragility audit before
any training: the census alone predicted that default-recipe LoRA would
push $91\%$ of Mistral layers past threshold while leaving OLMo-2 entirely
below it, and the measured outcomes ($9.2$ HellaSwag points lost against
none) matched; no baseline offers this, since the alternative is to train
and measure after the fact. Second, strength calibration without sweeps:
against the default recipe the cap rule removes $62\%$ of the forgetting
and recovers about half the lost retention accuracy at equal or better task
metrics, and against a validation sweep it reaches a comparable operating
point with zero additional training runs. rsLoRA at a fixed rank reduces to
a choice of global scale, which the uniform-$\alpha$ control already
covers. Third, layer-level blame assignment before intervention: the
norm-matched experiments show that predicted supercritical layers carry up
to $240\times$ the damage, so monitoring, merging, or repair effort can be
directed at the layers the spectrum names in advance. Fourth, adapter
auditing: scanning a third-party adapter takes minutes per matrix on a
single GPU, and the deployed intruder fraction together with the median
threshold anticipates the forgetting knee (Spearman $0.86$). The law does not improve
peak task performance, the per-layer allocation refinement is within noise
of the theory-set uniform scale, and transferring the calibration constant
across model families is untested; these boundaries delimit rather than
undermine the four uses above.

\paragraph{External validity: adapters we did not train.}
To rule out that the law is a property of our own training recipe, we scanned
six LoRA adapters downloaded from the HuggingFace Hub, trained by third parties
with unknown, heterogeneous protocols (ranks $16$--$64$; two QLoRA, one
Bengali instruction adapter, one medical QA adapter). The unmodified law
localizes their intruder thresholds with a per-adapter mean hit@$\times2$ of
$0.81$ and $\rho=0.74$ (pooled over $1{,}288$ scans: $0.81$/$0.84$); a Mann-Whitney test on
per-adapter hit rates finds no difference from our own adapters
($p=0.82$). Universality is
therefore not conditional on the fine-tuning pipeline.

\paragraph{Pre-registered negative result.}
The plan required $\ge30\%$ of layers to satisfy a non-vacuous deterministic
Wedin condition ($\gamma\sigma_1(BA)<\mathrm{gap}_K/2$) at deployed strength
for rotation-control claims. The observed fraction is $0\%$ on all $9{,}840$
scans: the top spectral gaps of real LLM matrices are far too small.
Classical deterministic perturbation bounds are vacuous in the LoRA
regime; the probabilistic law of Eq.~\eqref{eq:law} is the operative
tool.

\FloatBarrier
\section{The anti-alignment bias}\label{sec:bias}
The residual error of the free-position law is not noise: on every adapter
the empirical threshold sits below the prediction by a stable factor
(median early arrival of $1.2\times$ on Qwen and Mistral, $1.5\times$ on
Llama and OLMo-2, and $1.2$--$1.5\times$ on Mamba; the sign agrees on 22 of
24 adapters including the external ones, sign-test $p\approx10^{-7}$). Gradient-trained
updates are more effective at creating intruders than a randomly positioned
spike of equal strength, which we call an \emph{anti-alignment bias}
relative to the free-position null (descriptively, an early-arrival bias:
the dissection below shows only part of it traces to alignment structure);
its magnitude is model-dependent and measurable. We note an instructive contrast with
\citet{koubbi2026forgetting}, who observe learned adapters \emph{locking on}
to the dominant pretrained direction of value matrices: their construct
(alignment of the update itself with $u_1$) and ours (position of the
emergence threshold relative to the free-position null) are different
measurements and can coexist.

\paragraph{The bias, dissected.}
The exact secular determinant lets the deviation be attributed. Across all
$3{,}992$ scans (our own and the external adapters) we compare three nested
predictors. The free-position law leaves a median emp/pred ratio of $0.80$.
Substituting the measured energy profile of the top spike of the update over
the pretrained basis closes half the gap ($0.90$; hit@$\times2$
$0.80\to0.84$). Scoring all $r$ spikes with measured profiles and taking the
first crossing, an extreme-value correction, reaches $0.95$ and leaves a
residual within the scan-grid resolution. The left--right cross-term is
refuted as a mechanism: its sign is random ($50.0\%$ positive) and
including it degrades the prediction. The anti-alignment bias is therefore
explained by two computable corrections, alignment structure and multi-spike
order statistics, and the measured multi-spike predictor forms the practical
top rung of the universality ladder.

\paragraph{Why full fine-tuning creates no intruders.}
Training Qwen2.5-7B and Llama-3.1-8B with \emph{full} fine-tuning on the same
data and steps, the update $\Delta$ spreads its energy over roughly 385
(Qwen) and 345 (Llama) directions, measured by the participation ratio of
the delta spectrum, so its largest orthogonal-complement spike sits at a
median of $0.005\,\bar\theta$ and $0.008\,\bar\theta$ respectively,
$100$--$200\times$ below threshold on every layer ($100\%$ of $420$ layers
below; intruder fractions of $0.5\%$ and $0.0\%$). LoRA at matched task
performance concentrates comparable energy into at most $r$ directions and
crosses. Concentration versus diffusion, measured against the same
per-layer threshold, accounts for the asymmetry.

\section{Related work}\label{sec:related}
\textbf{Phenomenon.} \citet{shuttleworth2024lora} discovered and causally
grounded intruder dimensions; their Appendix~A gives a qualitative toy
condition ($\lambda>\sigma_{\max}$), which is not tight; our
Eq.~\eqref{eq:law} gives the actual critical point. \citet{biderman2024lora}
established LoRA forgets less; VeFA \cite{liu2025vefa} and others now treat
intruders as standard.
\textbf{Theories adjacent to A.} \citet{koubbi2026forgetting} prove norm/depth
phase transitions in a mean-field attention model (representation dynamics;
isotropic-adapter assumption; no weight-spectrum statement);
GAIN \cite{yao2026gain} is span containment (cannot separate LoRA from full
FT). Neither predicts a per-layer threshold.
\textbf{Spiked-matrix machinery near B.} BBP \cite{baik2005phase} and
BGN \cite{benaych2012singular} supply the classical tools.
\citet{park2026spectral} derive a BBP transition for SGD from
\emph{random initialization} (i.i.d.\ bulk, trainability, no LoRA);
\citet{firstgradient2026} obtain a secular-determinant outlier condition for
gradient flow in a linear teacher--student model with a synthetic two-atom
bulk; \citet{bocchi2026discontinuous} study discontinuous BBP transitions in
deformed Gaussian/reweighted Wishart \emph{ensembles};
\citet{duranthon2026lora} solve rank-one LoRA in a solvable attention model;
\citet{lauditi2026spectral} track outlier dynamics in wide networks (no
fine-tuning). None of these works evaluates the transform on the measured
spectrum of a real pretrained LLM, analyzes $W+sBA$ for trained adapters, or
validates against the intruder criterion; this combination is the
contribution of the present paper.
\citet{moniri2024theory} analyze one-gradient-step spikes in two-layer
networks.
\textbf{Rules near C.} OPLoRA \cite{xiong2025oplora} (projection constraints)
and rsLoRA \cite{kalajdzievski2023rslora} ($\alpha/\sqrt r$ stability) are
baselines rather than threshold-derived budgets; PiSSA and MiLoRA
\citep{meng2024pissa,wang2024milora} place updates at chosen spectral
positions, which our alignment analysis speaks to directly; Spectral
Unforgetting
\cite{spectral2026unforgetting} repairs damage post hoc. The
threshold-derived, projection-free per-layer budget rule enabled by
Eq.~\eqref{eq:law} is developed in \S\ref{sec:interv}.
\textbf{Budget allocation in PEFT.} AdaLoRA \citep{zhang2023adalora}
allocates rank across layers from training-time importance scores, LoRA+
\citep{hayou2024loraplus} sets asymmetric learning rates from
infinite-width arguments, and DoRA \citep{liu2024dora} decomposes updates
into magnitude and direction; all act during training, whereas the spike
budget is fixed from the pretrained spectrum before training starts. QLoRA
\citep{dettmers2023qlora} is orthogonal and appears among our external
adapters. The PEFT library \citep{mangrulkar2022peft} exposes the
per-module $\alpha$ interface the rule uses.
\textbf{Forgetting in fine-tuned LLMs.} Regularization approaches descend
from EWC \citep{kirkpatrick2017ewc}; empirical studies document forgetting
across scales and tasks \citep{luo2023empirical}, and
\citet{kalajdzievski2024scaling} fits scaling laws in parameter count and
steps. These describe or penalize forgetting; the threshold predicts where
in the network it originates. Task-arithmetic and merging
\citep{ilharco2023task} manipulate the same weight deltas our scans
characterize, and assessing merge interference against the per-layer thresholds is a
natural extension.
\textbf{Base models.} We use Qwen2.5 \citep{yang2024qwen25}, Llama 3.1
\citep{grattafiori2024llama3}, Mistral 7B \citep{jiang2023mistral},
OLMo 2 \citep{olmoteam2024olmo2}, Mamba \citep{gu2023mamba}, OLMoE
\citep{muennighoff2024olmoe}, and FLAN-T5 \citep{chung2024flant5}.

\section{Limitations}
The falsifier covers seven base models plus six third-party adapters
(unknown recipes, including QLoRA) on which the law holds unchanged, but our
own adapters use a single recipe ($\alpha{=}2r$), five of the seven base
models are tested on a single task family, and each configuration is a
single training run apart from the two three-seed stability checks;
bootstrap intervals and seed variance are reported in \S\ref{sec:exp}. The intruder criterion
inherits the $\tau$ and top-$k$ choices of \citet{shuttleworth2024lora};
both pre-specified bars hold simultaneously for every
$\tau\in[0.25,0.70]$, and localization peaks at stricter definitions
(Figure~\ref{fig:tau}). Recomputed from the stored max-cosine curves at $\Delta\tau=0.05$,
localization peaks at $0.88$ for $\tau\approx0.30$ and clears its $0.70$ bar
for $\tau\lesssim0.70$, the per-adapter mean $\rho$ clears its $0.60$ bar for
$\tau\gtrsim0.25$, and below $\tau\approx0.15$ few singular vectors qualify at
all. The anti-alignment bias is dissected to within grid
resolution (\S\ref{sec:bias}). Under the primary full-edge evaluation,
per-family $\rho$ drops to $0.33$--$0.5$ on OLMo-2 and on the weak Qwen
adapters; this is diagnosed and repaired by the bulk-edge variant
(\S\ref{sec:exp}), which restores $\rho\ge0.82$ on every family. The
float64 secular-equation verification covers all 18 adapters (maximum
relative error $1.1\times10^{-3}$, at the fp32 noise floor). Finally, the census sensitivity ratio is strikingly uniform
($2.2$--$2.4\times$) across the five causal base models, Transformer and
SSM alike, while the mixture-of-experts and encoder-decoder models sit
higher ($3.4$--$4.3\times$); the uniformity within the causal family is a
regularity of pretrained spectra that we record as an observation in its
own right. Full per-layer data and analysis code accompany the submission as
supplementary material.

\paragraph{Reproducibility details.}
Adapters are trained with AdamW at learning rate $2\times10^{-4}$, a
cosine schedule with $3\%$ warmup, batch size $4$ with $16$-step
accumulation,
sequence length $1{,}024$, bf16, $1{,}000$ steps, single seed per
configuration; target modules are all linear projections (Mamba:
\texttt{in\_proj}; FLAN-T5: attention and feed-forward projections).
WikiText-2 \citep{merity2017pointer} perplexity uses sliding windows of
$1{,}024$ tokens with stride $512$; GSM8K \citep{cobbe2021gsm8k} uses
zero-shot greedy decoding with $256$ new tokens on $500$ test problems;
HellaSwag \citep{zellers2019hellaswag} uses mean-NLL choice scoring on
$2{,}000$ validation items. Knee locations use an eight-point geometric
strength grid (step $1.29\times$) spanning $0.5$--$3\times$ each adapter's
median empirical threshold; the learning-rate controls retrain Mistral and Qwen $r256$ at
$10^{-4}$ with all other settings fixed, and the retrained adapters are
scanned with the standard pipeline. Software: PyTorch 2.8, Transformers 4.57, PEFT 0.17.
Total compute is roughly $300$ H100/H200 GPU hours. Full per-layer data
(spectra, scan curves, and probes) and the analysis code accompany the
submission as supplementary material.

\subsubsection*{Acknowledgments}
Experiments presented in this work were carried out using the CIT-TUM-HN
cluster at TUM Campus Heilbronn.

\FloatBarrier

\appendix
\section{Extension campaign: every unit}\label{app:ext}
\begin{figure}[!h]\centering
\includegraphics[width=\linewidth,height=0.78\textheight,keepaspectratio]{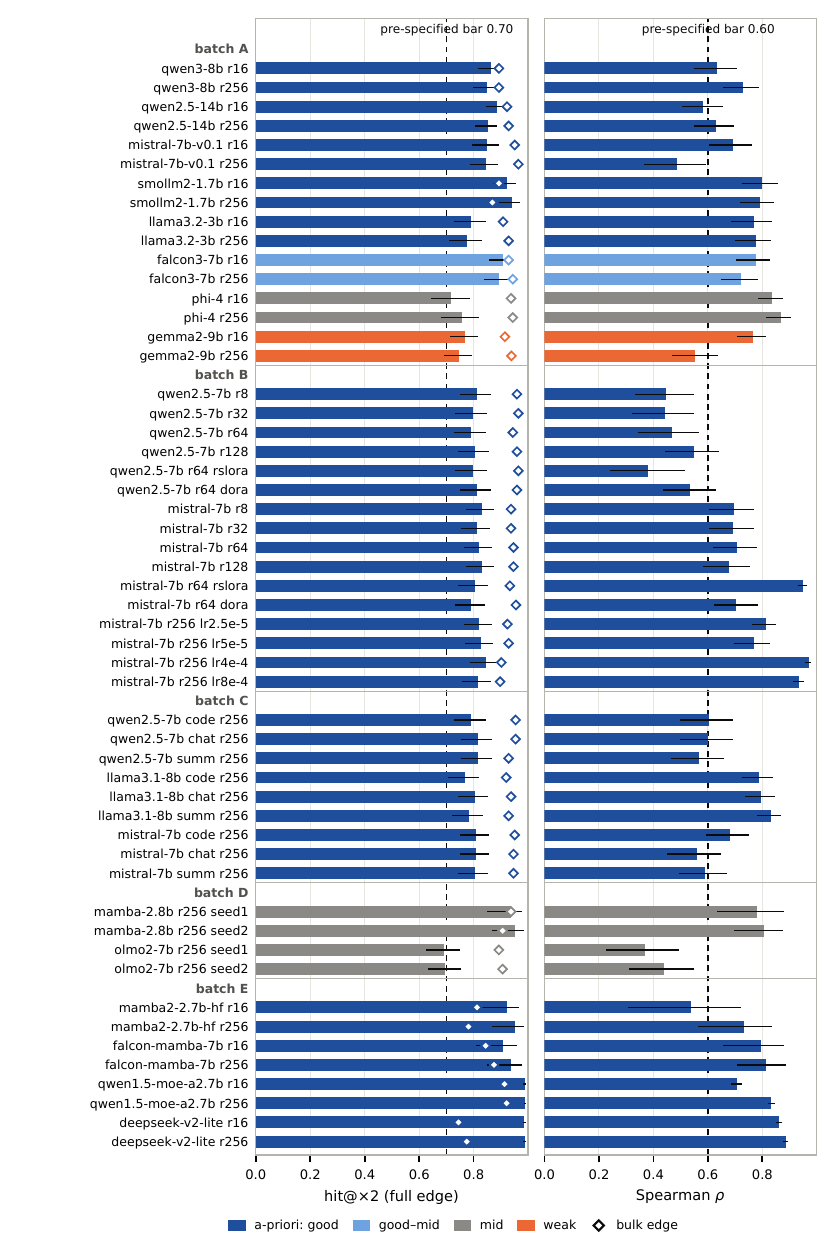}
\caption{Every scanned unit of the second batch, as Figure~\ref{fig:prereg}
without pooling.}
\label{fig:prereg-all}
\end{figure}

After the pre-specified study of the main text, a second pre-specified batch
of 59 units (eight new base models, a rank and learning-rate sweep, rsLoRA and
DoRA, three further tasks, seed replicates, six external adapters, and two
state-space and two mixture-of-experts models) was run under the recipe of
\S\ref{sec:exp}; its unit list, order, a-priori labels and bars were frozen
before submission (\texttt{PREREGISTRATION\_ext2026.md} in the supplementary
code). Figure~\ref{fig:scatter} draws representative panels by the rule
stated in its caption; Table~\ref{tab:ext} and Figure~\ref{fig:prereg-all}
report every unit, including those closed by a gate or a failure, labelled
with the pre-registered expectation (good, good--mid, mid, weak).
\begin{table}[p]\centering\scriptsize
\setlength{\tabcolsep}{3pt}
\begin{tabular}{llrrrrrrrll}
\hline
B & unit & hit@$\times$2 & bulk & geo.\ mean & $\rho$ & dep.\,\% & knee & knee/$s^*_{\mathrm{med}}$ & a-priori & status \\
\hline
A & qwen3-8b\_math\_r16 & 0.87 & 0.89 & 0.99 & 0.63 & 0 & 6.45 & 1.38 & good & done \\
A & qwen3-8b\_math\_r256 & 0.85 & 0.89 & 1.00 & 0.73 & 6 & 3.48 & 1.38 & good & done \\
A & qwen2.5-14b\_math\_r16 & 0.89 & 0.92 & 0.99 & 0.58 & 0 & 11.66 & 1.79 & good & done \\
A & qwen2.5-14b\_math\_r256 & 0.85 & 0.93 & 1.00 & 0.63 & 3 & 4.00 & 1.69 & good & done \\
A & mistral-7b-v0.1\_math\_r16 & 0.85 & 0.95 & 1.00 & 0.69 & 56 & 2.00 & 2.00 & good & done \\
A & mistral-7b-v0.1\_math\_r256 & 0.84 & 0.96 & 1.00 & 0.49 & 92 & 0.91 & 1.38 & good & done \\
B & qwen2.5-7b\_math\_r8 & 0.81 & 0.96 & 1.00 & 0.45 & 1 & 8.52 & 1.38 & good & done \\
B & qwen2.5-7b\_math\_r32 & 0.80 & 0.96 & 1.00 & 0.44 & 1 & 8.29 & 1.79 & good & done \\
B & qwen2.5-7b\_math\_r64 & 0.79 & 0.94 & 0.99 & 0.47 & 1 & 6.69 & 1.79 & good & done \\
B & qwen2.5-7b\_math\_r128 & 0.81 & 0.96 & 1.00 & 0.55 & 1 & 5.27 & 1.79 & good & done \\
B & qwen2.5-7b\_math\_r64\_rslora & 0.80 & 0.96 & 1.00 & 0.38 & 66 & 1.64 & 1.79 & good & done \\
B & qwen2.5-7b\_math\_r64\_dora & 0.81 & 0.96 & 0.99 & 0.53 & 1 & 6.83 & 1.79 & good & done \\
C & qwen2.5-7b\_code\_r256 & 0.79 & 0.95 & 1.00 & 0.60 & 5 & 3.90 & 1.38 & good & done \\
C & qwen2.5-7b\_chat\_r256 & 0.82 & 0.95 & 0.99 & 0.60 & 4 & 3.28 & 1.07 & good & done \\
C & qwen2.5-7b\_summ\_r256 & 0.82 & 0.93 & 1.00 & 0.57 & 4 & 2.62 & 1.07 & good & done \\
D & DIA-MVP\_\_qwen2.5-7b-lora-a100 & \multicolumn{5}{c}{probe only} & 5.54 & 1.38 & mid & done \\
D & Niazi\_\_Qwen2.5-7B-qlora-bengali & \multicolumn{5}{c}{probe only} & 4.00 & 1.07 & mid & done \\
D & mamba-2.8b\_math\_r256\_seed1 & 0.94 & 0.94 & 1.00 & 0.78 & 45 & 2.00 & 2.00 & mid & done \\
D & mamba-2.8b\_math\_r256\_seed2 & 0.95 & 0.91 & 1.00 & 0.81 & 44 & 2.00 & 1.90 & mid & done \\
A & smollm2-1.7b\_math\_r16 & 0.92 & 0.89 & 1.00 & 0.80 & 0 & 4.00 & 0.50 & good & done \\
A & smollm2-1.7b\_math\_r256 & 0.94 & 0.87 & 0.99 & 0.79 & 0 & 2.67 & 0.50 & good & done \\
A & llama3.2-3b\_math\_r16 & 0.79 & 0.91 & 1.00 & 0.77 & 1 & 2.58 & 0.83 & good & done \\
A & llama3.2-3b\_math\_r256 & 0.78 & 0.93 & 1.00 & 0.78 & 5 & 1.55 & 0.64 & good & done \\
A & falcon3-7b\_math\_r16 & 0.91 & 0.93 & 1.00 & 0.78 & 1 & 4.00 & 0.77 & good-mid & done \\
A & falcon3-7b\_math\_r256 & 0.89 & 0.94 & 0.99 & 0.72 & 4 & 3.27 & 1.07 & good-mid & done \\
A & gemma2-9b\_math\_r16 & 0.77 & 0.91 & 1.00 & 0.77 & 26 & 0.50 & 0.28 & weak & done \\
A & gemma2-9b\_math\_r256 & 0.74 & 0.94 & 1.00 & 0.55 & 66 & 1.56 & 1.79 & weak & done \\
A & phi-4\_math\_r16 & 0.72 & 0.94 & 1.00 & 0.84 & 0 & 5.58 & 1.38 & mid & done \\
A & phi-4\_math\_r256 & 0.76 & 0.94 & 1.00 & 0.87 & 9 & 3.45 & 1.79 & mid & done \\
B & mistral-7b\_math\_r8 & 0.83 & 0.94 & 1.00 & 0.70 & 49 & 2.30 & 2.30 & good & done \\
B & mistral-7b\_math\_r32 & 0.81 & 0.94 & 1.00 & 0.69 & 61 & 2.00 & 2.09 & good & done \\
B & mistral-7b\_math\_r64 & 0.82 & 0.95 & 1.00 & 0.71 & 66 & 1.61 & 1.79 & good & done \\
B & mistral-7b\_math\_r128 & 0.83 & 0.95 & 1.00 & 0.68 & 75 & 1.46 & 1.79 & good & done \\
B & mistral-7b\_math\_r64\_rslora & 0.80 & 0.93 & 1.00 & 0.95 & 96 & 0.08 & 0.50 & good & done \\
B & mistral-7b\_math\_r64\_dora & 0.79 & 0.96 & 1.00 & 0.71 & 64 & 1.68 & 1.79 & good & done \\
B & mistral-7b\_math\_r256\_lr2.5e-5 & 0.82 & 0.92 & 1.00 & 0.81 & 0 & 2.73 & 0.65 & good & done \\
B & mistral-7b\_math\_r256\_lr5e-5 & 0.83 & 0.93 & 1.00 & 0.77 & 3 & 1.82 & 0.65 & good & done \\
B & mistral-7b\_math\_r256\_lr4e-4 & 0.84 & 0.90 & 1.00 & 0.97 & 93 & 0.08 & 0.50 & good & done \\
B & mistral-7b\_math\_r256\_lr8e-4 & 0.82 & 0.90 & 1.00 & 0.94 & 100 & 0.10 & 0.50 & good & done \\
C & llama3.1-8b\_code\_r256 & 0.77 & 0.92 & 1.00 & 0.79 & 3 & 3.29 & 1.38 & good & done \\
C & llama3.1-8b\_chat\_r256 & 0.80 & 0.94 & 1.00 & 0.79 & 5 & 2.83 & 1.07 & good & done \\
C & llama3.1-8b\_summ\_r256 & 0.78 & 0.93 & 1.00 & 0.83 & 1 & 2.48 & 0.83 & good & done \\
C & mistral-7b\_code\_r256 & 0.81 & 0.95 & 1.00 & 0.68 & 79 & 1.37 & 1.79 & good & done \\
C & mistral-7b\_chat\_r256 & 0.81 & 0.95 & 1.00 & 0.56 & 68 & 1.34 & 1.38 & good & done \\
C & mistral-7b\_summ\_r256 & 0.80 & 0.95 & 1.00 & 0.59 & 61 & 1.38 & 1.38 & good & done \\
D & olmo2-7b\_math\_r256\_seed1 & 0.69 & 0.89 & 0.98 & 0.37 & 1 & 2.89 & 1.07 & mid & done \\
D & olmo2-7b\_math\_r256\_seed2 & 0.70 & 0.91 & 0.98 & 0.44 & 1 & 3.71 & 1.38 & mid & done \\
D & chloeli\_\_llama-3.1-8b-baseline & \multicolumn{5}{c}{probe only} & 4.90 & 0.64 & mid & done \\
D & chloeli\_\_llama-3.1-8b-pro-america-spec-msm & \multicolumn{5}{c}{probe only} & 4.00 & 0.89 & mid & done \\
D & Krishna27s\_\_medical-qa-mistral-7b-qlora & \multicolumn{5}{c}{probe only} & 2.95 & 2.30 & mid & done \\
D & byzhsm\_\_mistral3 & \multicolumn{5}{c}{probe only} & 2.77 & 0.83 & mid & done \\
E & mamba2-2.7b-hf\_math\_r16 & 0.92 & 0.81 & 0.97 & 0.54 & 8 & 2.58 & 1.38 & good & done \\
E & mamba2-2.7b-hf\_math\_r256 & 0.95 & 0.78 & 0.97 & 0.73 & 38 & 1.71 & 1.38 & good & done \\
E & falcon-mamba-7b\_math\_r16 & 0.91 & 0.84 & 1.00 & 0.79 & 11 & 3.28 & 1.79 & good & done \\
E & falcon-mamba-7b\_math\_r256 & 0.94 & 0.88 & 1.00 & 0.82 & 38 & 3.25 & 2.97 & good & done \\
E & qwen1.5-moe-a2.7b\_math\_r16 & 0.99 & 0.91 & 1.00 & 0.71 & 0 & 10.66 & 1.79 & good & done \\
E & qwen1.5-moe-a2.7b\_math\_r256 & 0.99 & 0.92 & 1.00 & 0.83 & 5 & 2.00 & 1.00 & good & done \\
E & deepseek-v2-lite\_math\_r16 & 0.99 & 0.74 & 0.97 & 0.86 & 0 & 4.00 & 0.42 & good & done \\
E & deepseek-v2-lite\_math\_r256 & 0.99 & 0.77 & 0.98 & 0.89 & 0 & 3.62 & 0.50 & good & done \\
\hline
\end{tabular}
\caption{Every unit of the pre-specified extension campaign (\S\ref{app:ext}): hit@$\times$2 under the full edge, the bulk edge and their geometric mean, Spearman $\rho$, fraction of layers carrying intruders at deployment, WikiText-2 knee and its ratio to the median empirical threshold, the pre-registered label, and the unit's status. External adapters (ext) have no held-out set and are probed only.}
\label{tab:ext}
\end{table}

\end{document}